\documentclass[10pt,twocolumn,letterpaper]{article}

\usepackage{cvpr}
\usepackage{times}
\usepackage{epsfig}
\usepackage{graphicx}
\usepackage{amsmath}
\usepackage{amssymb}
\usepackage{textcomp,booktabs}
\usepackage[usenames,dvipsnames]{color}
\usepackage{colortbl}
\definecolor{mygray}{gray}{.9}
\definecolor{mypink}{rgb}{.99,.91,.95}
\definecolor{mycyan}{cmyk}{.3,0,0,0}

\usepackage[pagebackref=true,breaklinks=true,letterpaper=true,colorlinks,bookmarks=false]{hyperref}

\cvprfinalcopy 

\def\cvprPaperID{373} 

\ifcvprfinal\fi
\begin{document}

\title{Borrowing Treasures from the Wealthy: Deep Transfer Learning through Selective Joint Fine-Tuning}

\author{Weifeng Ge \hspace{1.2in} Yizhou Yu\\
Department of Computer Science, The University of Hong Kong
}

\maketitle

\begin{abstract}
Deep neural networks require a large amount of labeled training data during supervised learning. However, collecting and labeling so much data might be infeasible in many cases. 
In this paper, we introduce a deep transfer learning scheme, called selective joint fine-tuning, for improving the performance of deep learning tasks with insufficient training data. In this scheme, a target learning task with insufficient training data is carried out simultaneously with another source learning task with abundant training data. However, the source learning task does not use all existing training data.
Our core idea is to identify and use a subset of training images from the original source learning task whose low-level characteristics are similar to those from the target learning task, and jointly fine-tune shared convolutional layers for both tasks. Specifically,
we compute descriptors from linear or nonlinear filter bank responses on training images from both tasks, and use such descriptors to search for a desired subset of training samples for the source learning task.


Experiments demonstrate that our deep transfer learning scheme achieves state-of-the-art performance on multiple visual classification tasks with insufficient training data for deep learning. Such tasks include Caltech 256, MIT Indoor 67, and fine-grained classification problems (Oxford Flowers 102 and Stanford Dogs 120). In comparison to fine-tuning without a source domain, the proposed method can improve the classification accuracy by 2\% - 10\% using a single model. Codes and models are available at \url{https://github.com/ZYYSzj/Selective-Joint-Fine-tuning}.
\end{abstract}

\section{Introduction}\label{sec:intro}
Convolutional neural networks (CNNs) have become deeper and larger to pursue increasingly better performance on classification and recognition tasks~\cite{AlexNet,VGGnet,GoogleNet,he2015deep,HDCNN2015}. Looking at the successes of deep learning in computer vison, we find that a large amount of training or pre-training data is essential in training deep neural networks. Large-scale image datasets, such as the ImageNet ILSVRC dataset~\cite{russakovsky2015imagenet}, Places~\cite{zhou2014learning}, and MS COCO~\cite{lin2014microsoft}, have led to a series of breakthroughs in visual recognition, including image classification~\cite{lin2015bilinear}, object detection~\cite{girshick2014rich}, and semantic segmentation~\cite{long2015fully}. Many other related visual tasks have benefited from these breakthroughs.


Nonetheless, researchers face a dilemma when using deep convolutional neural networks to perform visual tasks that do not have sufficient training data. Training a deep network with insufficient data might even give rise to inferior performance in comparison to traditional classifiers fed with handcrafted features. Fine-grained classification problems, such as Oxford Flowers 102 \cite{nilsback2008automated} and Stanford Dogs 120 \cite{khosla2011novel}, are such examples.
The number of training samples in these datasets is far from being enough for training large-scale deep neural networks, and the networks would become overfit quickly.

Solving the overfitting problem for deep convolutional neural networks on learning tasks without sufficient training data is challenging~\cite{srivastava2014dropout}.
Transfer learning techniques that apply knowledge learnt from one task to other related tasks have been proven helpful~\cite{pan2010survey}. In the context of deep learning, fine-tuning a deep network pre-trained on the ImageNet or Places dataset is a common strategy to learn task-specific deep features.
This strategy is considered a simple transfer learning technique for deep learning. However, since the ratio between the number of learnable parameters and the number of training samples still remains the same, fine-tuning needs to be terminated after a relatively small number of iterations; otherwise, overfitting still occurs.

In this paper, we attempt to tackle the problem of training deep neural networks for learning tasks that have insufficient training data. We adopt the source-target joint training methodology~\cite{xue2007multi} when fine-tuning deep neural networks. The original learning task without sufficient training data is called the target learning task, $\boldsymbol{\mathcal{T}}_{t}$. To boost its performance, the target learning task is teamed up with another learning task with rich training data. The latter is called the source learning task, $\boldsymbol{\mathcal{T}}_{s}$. Suppose the source learning task has a large-scale training set $\boldsymbol{\mathcal{D}}_{s}$,
and the target learning task has a small-scale training set $\boldsymbol{\mathcal{D}}_{t}$.
Since the target learning task is likely a specialized task, we envisage the image signals in its dataset possess certain unique low-level characteristics (e.g. fur textures in Stanford Dogs 120 \cite{khosla2011novel}), and the learned kernels in the convolutional layers of a deep network need to grasp such characteristics in order to generate highly discriminative features. Thus supplying sufficient training images with similar low-level characteristics becomes the most important mission of the source learning task. Our core idea is to identify a subset of training images from $\boldsymbol{\mathcal{D}}_{s}$ whose low-level characteristics are similar to those from $\boldsymbol{\mathcal{D}}_{t}$, and then jointly fine-tune a shared set of convolutional layers for both source and target learning tasks. The source learning task is fine-tuned using the selected training images only. Hence, this process is called {\em selective joint fine-tuning}. The rationale behind this is that the unique low-level characteristics of the images from $\boldsymbol{\mathcal{D}}_{t}$ might be overwhelmed if all images from $\boldsymbol{\mathcal{D}}_{s}$ were taken as training samples for the source learning task.

How do we select images from $\boldsymbol{\mathcal{D}}_{s}$ that share similar low-level characteristics as those from $\boldsymbol{\mathcal{D}}_{t}$? Since kernels followed with nonlinear activation in a deep convolutional neural network (CNN) are actually nonlinear spatial filters, to find sufficient data for training high-quality kernels, we use the responses from existing linear or nonlinear filter banks to define similarity in low-level characteristics. Gabor filters~\cite{manjunath1996texture} form an example of a linear filter bank, and the complete set of kernels from certain layers of a pre-trained CNN form an example of a nonlinear filter bank. We use histograms of filter bank responses as image descriptors to search for images with similar low-level characteristics.

The motivation behind selecting images according to their low-level characteristics is two fold. First, low-level characteristics are extracted by kernels in the lower convolutional layers of a deep network. These lower convolutional layers form the foundation of an entire network, and the quality of features extracted by these layers determines the quality of features at higher levels of the deep network. Sufficient training images sharing similar low-level characteristics could strength the kernels in these layers. Second, images with similar low-level characteristics could have very different high-level semantic contents. Therefore, searching for images using low-level characteristics has less restrictions and can return much more training images than using high-level semantic contents.

The above source-target selective joint fine-tuning scheme is expected to benefit the target learning task in two different ways. First, since convolutional layers are shared between the two learning tasks, the selected training samples for the source learning task prevent the deep network from overfitting quickly. Second, since the selected training samples for the source learning task share similar low-level characteristics as those from the target learning task, kernels in their shared convolutional layers can be trained more robustly to generate highly discriminative features for the target learning task.

The proposed source-target selective joint fine-tuning scheme is easy to implement. Experimental results demonstrate state-of-the-art performance on multiple visual classification tasks with much less training samples than what is required by recent deep learning architectures. These visual classification tasks include fine-grained classification on Stanford Dogs 120~\cite{khosla2011novel} and Oxford Flowers 102~\cite{nilsback2008automated}, image classification on Caltech 256~\cite{griffin2007caltech}, and scene classification on MIT Indoor 67~\cite{quattoni2009recognizing}.

In summary, this paper has the following contributions:
{\flushleft $\bullet$} We introduce a new deep transfer learning scheme, called selective joint fine-tuning, for improving the performance of deep learning tasks with insufficient training data. 
It is an important step forward in the context of the widely adopted strategy of fine-tuning a pre-trained deep neural network.
{\flushleft $\bullet$} We develop a novel pipeline for implementing this deep transfer learning scheme. Specifically, we compute descriptors from linear or nonlinear filter bank responses on training images from both tasks, and use such descriptors to search for a desired subset of training samples for the source learning task.
{\flushleft $\bullet$} Experiments demonstrate that our deep transfer learning scheme achieves state-of-the-art performance on multiple visual classification tasks with insufficient training data for deep learning.

\section{Related Work}

\begin{figure*}[t]
  \centering
  \includegraphics[width=1.0\linewidth]{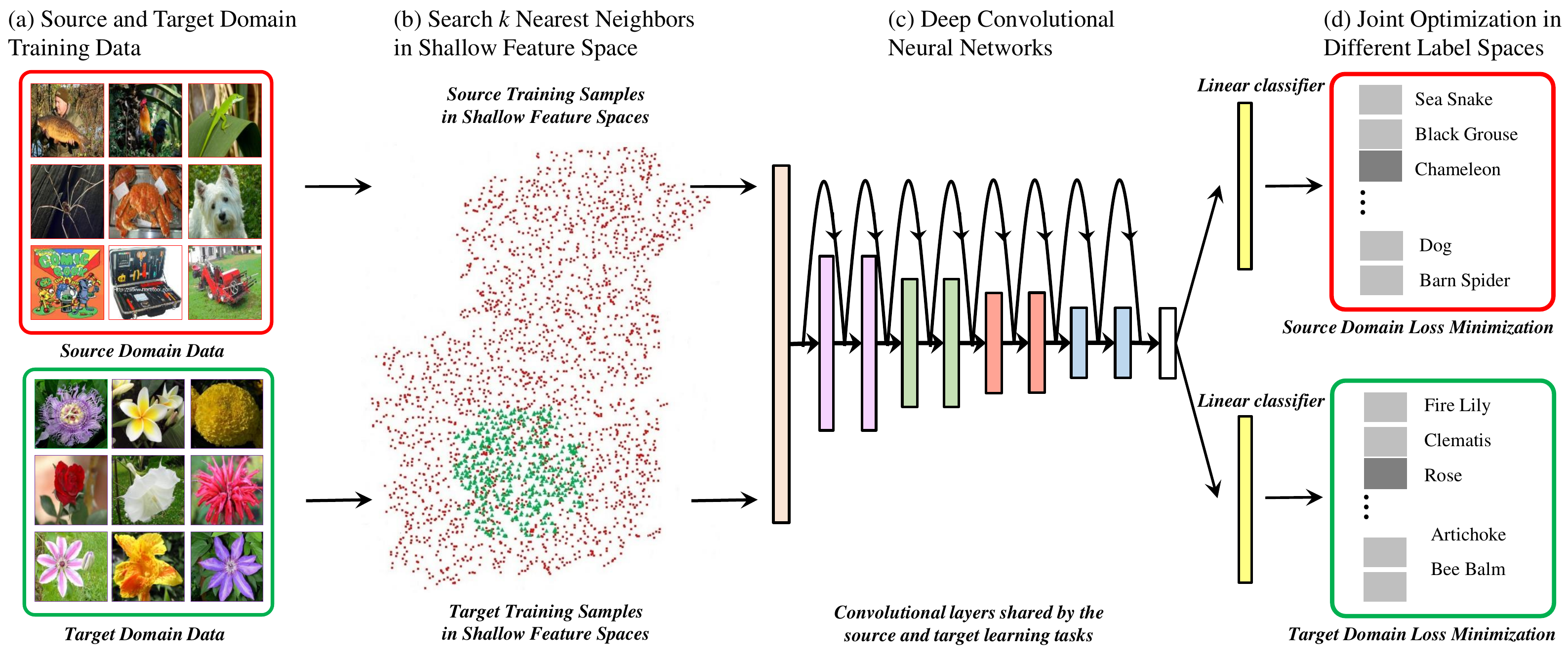}
  \caption{Pipeline of the proposed selective joint fine-tuning. From left to right: (a) Datasets in the source domain and the target domain. (b) Select nearest neighbors of each target domain training sample in the source domain via a low-level feature space. (c) Deep convolutional neural network initialized with weights pre-trained on ImageNet or Places. (d) Jointly optimize the source and target cost functions in their own label spaces.\vspace{-0mm}}
  \label{Fig:Transfer Learning Pipeline}
\end{figure*}

\noindent\textbf{Multi-Task Learning.} Multi-task learning (MTL) obtains shared feature representations or classifiers for related tasks~\cite{caruana1998multitask}. In comparison to learning individual tasks independently, features and classifiers learned with MTL often have better generalization capability.
In deep learning, faster RCNN~\cite{ren2015faster} jointly learns object locations and labels using shared convolutional layers but different loss functions for these two tasks.
In \cite{eigen2015predicting}, the same multi-scale convolutional architecture was used to predict depth, surface normals and semantic labels. This indicates that convolutional neural networks can be adapted to different tasks easily.
While previous work~\cite{evgeniou2007multi,ren2015faster} attempts to find a shared feature space that benefits multiple learning tasks, the proposed joint training scheme in this paper focuses on learning a shared feature space that improves the performance of the target learning task only.
\vspace{-0mm}

\noindent\textbf{Feature Extraction and Fine-tuning.} Off-the-shelf CNN features~\cite{sharif2014cnn,donahue2014decaf} have been proven to be powerful in various computer vision problems. Pre-training convolutional neural networks on ImageNet \cite{russakovsky2015imagenet} or Places \cite{zhou2014learning} has been the standard practice for other vision problems. However, features learnt in pre-trained models are not tailored for the target learning task. Fine-tuning pre-trained models \cite{girshick2014rich} has become a commonly used method to learn task-specific features. The transfer ability of different convolutional layers in CNNs has been investigated in \cite{yosinski2014transferable}. However, for tasks that do not have sufficient training data, overfitting occurs quickly during fine-tuning.
The proposed pipeline in this paper not only alleviates overfitting, but also attempts to find a more discriminative feature space for the target learning task.
\vspace{-0mm}

\noindent\textbf{Transfer Learning.} Different from MTL, transfer learning (or domain adaptation)~\cite{pan2010survey} applies knowledge learnt in one domain to other related tasks.
Domain adaptation algorithms can be divided into three categories, including instance adaption~\cite{huang2006correcting,azizpour2015generic}, feature adaption~\cite{long2015learning,tzeng2015simultaneous}, and model adaption~\cite{duan2012visual}.
Hong {\em et al.}~\cite{hong2016learning} transferred rich semantic information from source categories to target categories via the attention model.
Tzeng {\em et al.}~\cite{tzeng2015simultaneous} performed feature adaptation using a shared convolutional neural network by transferring the class relationship in the source domain to the target domain. To make our pipeline more flexible, this paper does not assume the source and target label spaces are the same as in \cite{tzeng2015simultaneous}. Different from the work in \cite{azizpour2015generic} which randomly resamples training classes or images in the source domain, this paper conducts a special type of transfer learning by selecting source training samples that are nearest neighbors of samples in the target domain in the space of certain low-level image descriptor.

Krause {\em et al.}~\cite{krause2015unreasonable} directly performed Google image search using keywords associated with categories from the target domain, and download a noisy collection of images to form a training set.
In our method, we search for nearest neighbors in a large-scale labeled dataset using low-level features instead of high-level semantic information. It has been shown in \cite{mahendran2015understanding} that low-level features computed in the bottom layers of a CNN encode very rich information, which can completely reconstruct the original image. Our experimental results show that nearest neighbor search using low-level features can outperform that using high-level semantic information as in \cite{krause2015unreasonable}.

\section{Selective Joint Fine-tuning}

\subsection{Overview}

Fig. \ref{Fig:Transfer Learning Pipeline} shows the overall pipeline for our proposed source-target selective joint fine-tuning scheme. Given a target learning task $\boldsymbol{\mathcal{T}}_{t}$ that has insufficient training data, we perform selective joint fine-tuning as follows. The entire training dataset associated with the target learning task is called the target domain. The source domain is defined similarly.
\vspace{-3mm}
\paragraph{Source Domain}: The minimum requirement is that the number of images in the source domain, $\boldsymbol{\mathcal{D}}_{s} = \big\{ \big(\boldsymbol{x}^s_i,y^s_i\big) \big\}^{n_{s}}_{i=1}$, should be large enough to train a deep convolutional neural network from scratch. Ideally, these training images should present diversified low-level characteristics. That is, running a filter bank on them give rise to as diversified responses as possible. There exist a few large-scale visual recognition datasets that can serve as the source domain, including ImageNet ILSVRC dataset~\cite{russakovsky2015imagenet}, Places~\cite{zhou2014learning}, and MS COCO~\cite{lin2014microsoft}.
\vspace{-3mm}
\paragraph{Source Domain Training Images}: In our selective joint fine-tuning, we do not use all images in the source domain as training images. Instead, for each image from the target domain, we search a certain number of images with similar low-level characteristics from the source domain. Only images returned from these searches are used as training images for the source learning task in selective joint fine-tuning. We apply a filter bank to all images in both source domain and target domain. Histograms of filter bank responses are used as image descriptors during search. We associate an adaptive number of source domain images with each target domain image. Hard training samples in the target domain might be associated with a larger number of source domain images. Two filter banks are used in our experiments. One is the Gabor filter bank, and the other consists of kernels in the convolutional layers of AlexNet pre-trained on ImageNet~\cite{AlexNet}.
\vspace{-4mm}
\paragraph{CNN Architecture}: Almost any existing deep convolutional neural network,
such as AlexNet~\cite{AlexNet}, VGGNet~\cite{VGGnet}, and ResidualNet \cite{he2015deep}, can be used in our selective joint fine-tuning. We use the 152-layer residual network with identity mappings~\cite{he2016identity} as the CNN architecture in our experiments. The entire residual network is shared by the source and target learning tasks. An extra output layer is added on top of the residual network for each of the two learning tasks. This output layer is not shared because the two learning tasks may not share the same label space. The residual network is pre-trained either on ImageNet or Places.
\vspace{-4mm}
\paragraph{Source-Target Joint Fine-tuning}: Each task uses its own cost function during selective joint fine-tuning, and every training image only contributes to the cost function corresponding to the domain it comes from. The source domain images selected by the aforementioned searches are used as training images for the source learning task only while the entire target domain is used as the training set for the target learning task only. Since the residual network (with all its convolutional layers) is shared by these two learning tasks, it is fine-tuned by both training sets. And the output layers on top of the residual network are fine-tuned by its corresponding training set only. Thus we conduct end-to-end joint fine-tuning to minimize the original loss functions of the source learning task and the target learning task simultaneously.
\vspace{-0mm}

\subsection{Similar Image Search}
There is a unique step in our pipeline. For each image from the target domain, we search a certain number of images with similar low-level characteristics from the source domain. Only images returned from these searches are used as training images for the source learning task in selective joint fine-tuning. We elaborate this image search step below.
\vspace{-4mm}

\paragraph{Filter Bank} We use the responses to a filter bank to describe the low-level characteristics of an image. The first filter bank we use is the Gabor filter bank. Gabor filters are commonly used for feature description, especially texture description~\cite{manjunath1996texture}. Gabor filter responses are powerful low-level features for image and pattern analysis. We use the parameter setting in \cite{manjunath1996texture} as a reference. For each of the real and imaginary parts, we use 24 convolutional kernels with 4 scales and 6 orientations. Thus there are 48 Gabor filters in total.

Kernels in a deep convolutional neural network are actually spatial filters. When there is nonlinear activation following a kernel, the combination of the kernel and nonlinear activation is essentially a nonlinear filter. A deep CNN can extract low/middle/high level features at different convolutional layers~\cite{yosinski2014transferable}. Convolutional layers close to the input data focus on extract low-level features while those further away from the input extract middle- and high-level features. In fact, a subset of the kernels in the first convolutional layer of AlexNet trained on ImageNet exhibit oriented stripes, similar to Gabor filters~\cite{AlexNet}. When trained on a large-scale diverse dataset, such as ImageNet, such kernels can be used for describing generic low-level image characteristics. In practice, we use all kernels (and their following nonlinear activation) from the first and second convolutional layers of AlexNet pre-trained on ImageNet as our second choice of a filter bank.
\vspace{-4mm}

\paragraph{Image Descriptor}
Let $\mathbb{C}_i(m,n)$ denote the response map to the $i$-th convolutional kernel or Gabor filter in our filter bank,
and $\boldsymbol{\phi}_i$ its histogram.
To obtain more discriminative histogram features, we first obtain the upper bound $h_i^u$ and lower bound $h_i^l$ of the $i$-th response map by scanning the entire target domain $\boldsymbol{\mathcal{D}}_{t}$. Then the interval ${h_i^l,h_i^u}$ is divided into a set of small bins.
We adaptively set the width of every histogram bin so that each of them contains a roughly equal percentage of pixels. In this manner, we can avoid a large percentage of pixels falling into the same bin.
We concatenate the histograms of all filter response maps to form a feature vector, $\boldsymbol{\phi}^k=\big\{\boldsymbol{\phi}_1,\boldsymbol{\phi}_2,¡­ ,\boldsymbol{\phi}_D\big\}$, for image $\boldsymbol{x}_k$.
\vspace{-4mm}

\paragraph{Nearest Neighbor Ranking}
Given the histogram-based descriptor of a training image $\boldsymbol{x}^t_i$ in the target domain, we search for its nearest neighbors in the source domain $\boldsymbol{\mathcal{D}}_{s}$. Note that the number of kernels in different convolutional layers of AlexNet might be different. To ensure equal weighting among different convolutional layers during nearest neighbor search, each histogram of kernel responses is normalized by the total number of kernels in the corresponding layer. Thus the distance between the descriptor of a source image $\boldsymbol{x}^s_j$ and that of a target image $\boldsymbol{x}^t_i$ is computed as follows.
    \begin{equation*}
    \begin{aligned}
      \mathcal{H} \big(\boldsymbol{x}^t_i,\boldsymbol{x}^s_j\big) = \sum_{h=1}^{D} w_h [\kappa(\boldsymbol{\phi}_h^{i,t},\boldsymbol{\phi}_h^{j,s})+ \kappa(\boldsymbol{\phi}_h^{j,s},\boldsymbol{\phi}_h^{i,t})],
    \end{aligned}
    \label{eq:nndist}
    \end{equation*}
where $w_h = 1/N_h$, $N_h$ is the number of convolutional kernels in the corresponding layer, $\boldsymbol{\phi}_h^{i,t}$ and $\boldsymbol{\phi}_h^{j,s}$ are the $h$-th histogram for images $x_i^t$ and $x_j^s$, and $\kappa(\cdot,\cdot)$ is the KL-divergence.
\vspace{-4mm}

\paragraph{Hard Samples in the Target Domain}
The labels of training samples in the target domain have varying degrees of difficulty to satisfy. Intuitively, we would like to seek extra help for those hard training samples in the target domain by searching for more and more nearest neighbors in the source domain.
We propose an iterative scheme for this purpose. We calculate the information entropy to measure the classification uncertainty of training samples in the target domain after the $m$-th iteration as follows.
    \begin{equation*}
    \begin{aligned}
           \mathcal{H}^m_i = - \sum_{c=1}^{C} p^m_{i,c}\log(p^m_{i,c}),
    \end{aligned}
    \label{eq:entropy}
    \end{equation*}
where $C$ is the number of classes, $p^m_{i,c}$ is the probability that the $i$-th training sample belongs to the $c$-th class after a softmax layer in the $m$-th iteration.

Training samples that have high classification uncertainty are considered hard training samples. In the next iteration, we increase the number of nearest neighbors of the hard training samples as in Eq. (\ref{eq:nnsearch}), and continue fine-tuning the model trained in the current iteration. For a training sample $\boldsymbol{x}^t_i$ in the target domain, the number of its nearest neighbors in the next iteration is defined as follows.
\begin{equation*}\label{eq:nnsearch}
\mathcal{K}^{m+1}_i=
\left\{
             \begin{array}{lr}
             \mathcal{K}^m_i + {\sigma}_0, & \widehat{y}^t_i \neq y^t_i \\
             \mathcal{K}^m_i + {\sigma}_1, & \widehat{y}^t_i = y^t_i \quad and \quad \mathcal{H}^m_i \geq \delta  \\
             \mathcal{K}^m_i,              & \widehat{y}^t_i = y^t_i \quad and \quad \mathcal{H}^m_i < \delta
             \end{array}
\right.
\eqno(6)
\end{equation*}
where ${\sigma}_0$, ${\sigma}_1$ and $\delta$ are constants, $\widehat{y}^t_i$ is predicted label of $\boldsymbol{x}^t_i$, and $\mathcal{K}_i^m$ is the number of nearest neighbors in the $m$-th iteration. By changing the number of nearest neighbors for samples in the target domain, the subset of the source domain used as training data evolves over iterations, which in turn gradually changes the feature representation learned in the deep network. In the above equation, we typically set $\delta= 0.1$, ${\sigma}_0 = 4\mathcal{K}_0$ and ${\sigma}_1 = 2\mathcal{K}_0$, where $\mathcal{K}_0$ is the initial number of nearest neighbors for all samples in the target domain. In our experiments, we stop after five iterations.

In Table \ref{tab:FilterBank}, we compare the effectiveness of Gabor filters and various combinations of kernels from AlexNet in our selective joint fine-tuning. In this experiment, we use the 50-layer residual network~\cite{he2015deep} with half of the convolutional kernels in the original architecture.

\begin{table}[h]\small
\setlength{\abovecaptionskip}{10pt}
\setlength{\belowcaptionskip}{-10pt}
\begin{center}
\begin{tabular}{@{}lcc@{}}
\toprule
Filter Bank                    & over all Accuracy(\%)        \\ \midrule
Conv1-Conv2 in AlexNet         & \textbf{89.59}               \\
Conv1-Conv5 in AlexNet         & 88.82                        \\
Conv4-Conv5 in AlexNet         & 88.48                        \\
Gabor Filters                  & 88.90                        \\
Fine-tuning w/o source domain  & 88.12                        \\
\bottomrule
\end{tabular}
\end{center}
\caption{A comparison of classification performance on Oxford Flowers 102 using various choices for the filter bank in selective joint fine-tuning.}
\label{tab:FilterBank}
\end{table}

\begin{figure*}[t]
  \centering
  \includegraphics[width=1.0\linewidth]{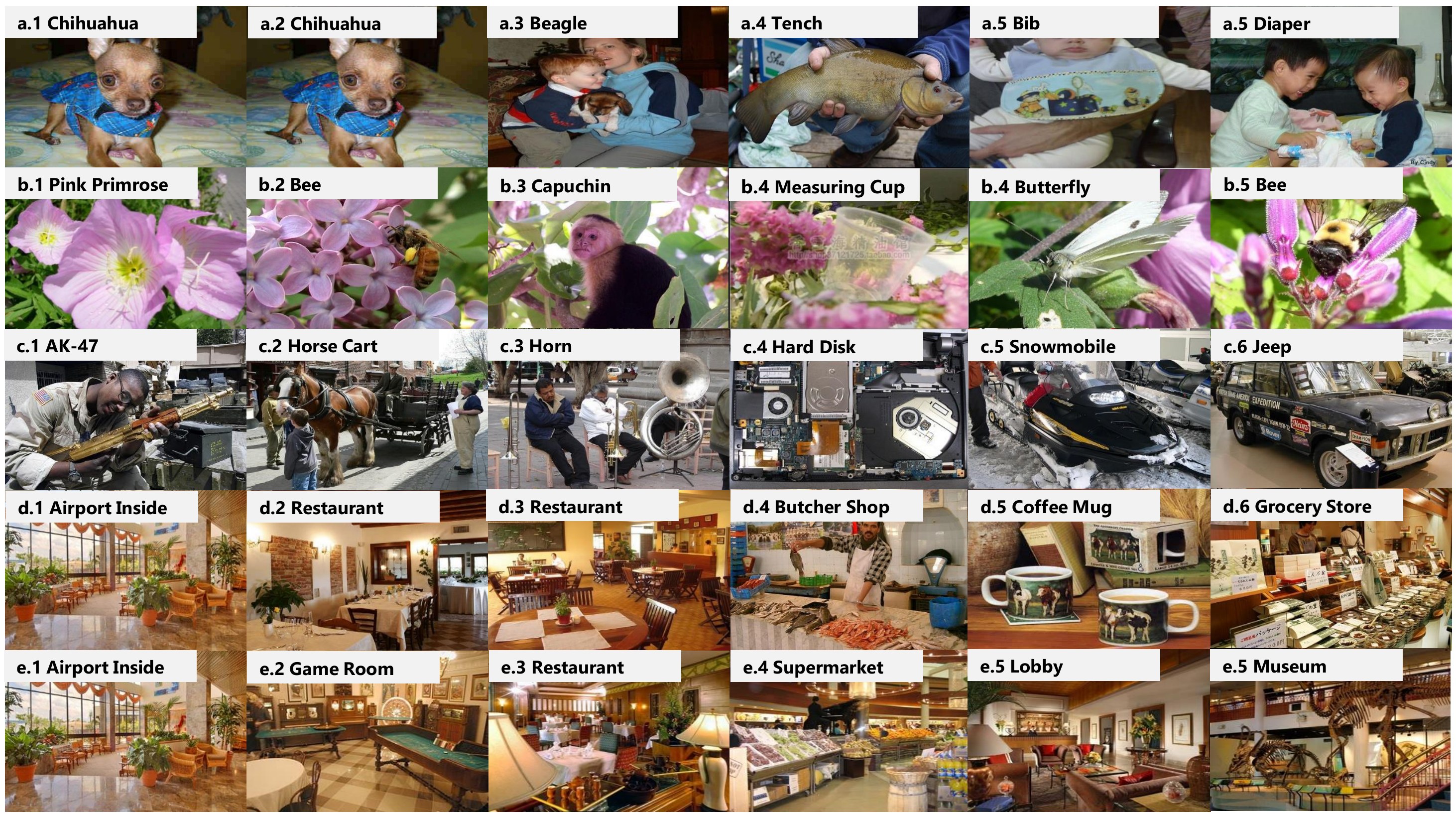}
  \caption{Images in the source domain that have similar low-level characteristics with the target images. The first column shows target images from Stanford Dogs 120~\cite{khosla2011novel}, Oxford Flowers 102~\cite{nilsback2008automated}, Caltech 256~\cite{griffin2007caltech}, and MIT Indoor 67~\cite{quattoni2009recognizing}. The following columns in rows (a)-(d) are the corresponding 1st, 10-th, 20-th, 30-th and 40-th nearest images in ImageNet (source domain). The following columns in row (e) are images retrieved from Places (source domain for MIT Indoor 67).\vspace{-0mm}}
  \label{Fig:Image Retrieval Results}
\end{figure*}

\section{Experiments}
\subsection{Implementation}
In all experiments, we use the 152-layer residual network with identity mappings~\cite{he2016identity} as the deep convolutional architecture, and conventional fine-tuning performed on a pre-trained network with the same architecture without using any source datasets as our {\em baseline}. Note that the network architecture we use is different from those used in most published methods for the datasets we run experiments on, and many existing methods adopt sophisticated parts models and feature encodings. The performance of such methods are still included in this paper to indicate that our simple holistic method without incorporating parts models and feature encodings is capable of achieving state-of-the-art performance.

We use the pre-trained model released in \cite{he2015deep} to initialize the residual network.
During selective joint fine-turning, source and target samples are mixed together in each mini-batch. Once the data has passed the average pooling layer in the residual network, we split the source and target samples, and send them to their corresponding softmax classifier layer respectively. Both the source and target classifiers are initialized randomly.

We run all our experiments on a TITAN X GPU with 12GB memory. All training data is augmented as in \cite{paulin2014transformation} first, and we follow the training and testing settings in \cite{he2015deep}.
Every mini-batch can include 20 224$\times$224 images using a modified implementation of the residual network. We include randomly chosen samples from the target domain in a mini-batch. Then for each of the chosen target sample, we further include one of its retrieved nearest neighbors from the source domain in the same mini-batch. We set the iter size to 10 for each iteration in Caffe \cite{jia2014caffe}. The momentum parameter is set to 0.9 and the weight decay is 0.0001 in SGD. During selective joint fine-tuning, the learning rate starts from 0.01 and is divided by 10 after every $2400-5000$ iterations in all the experiments. Most of the experiments can finish in 16000 iterations.

\subsection{Source Image Retrieval}
We use the ImageNet ILSVRC 2012 training set~\cite{russakovsky2015imagenet} as the source domain for Stanford Dogs~\cite{khosla2011novel}, Oxford Flowers~\cite{nilsback2008automated}, and Caltech 256~\cite{griffin2007caltech}, and the combination of the ImageNet and Places 205~\cite{zhou2014learning} training sets as the source domain for MIT Indoor 67~\cite{quattoni2009recognizing}. Fig. \ref{Fig:Image Retrieval Results} shows the retrieved 1-st, 10-th, 20-th, 30-th, and 40-th nearest neighbors from ImageNet~\cite{russakovsky2015imagenet} or Places~\cite{zhou2014learning}. It can be observed that corresponding source and target images share similar colors, local patterns and global structures.
Since low-level filter bank responses do not encode strong semantic information, the 50 nearest neighbors from a target domain include images from various and sometimes completely unrelated categories.

We find out experimentally that there should be at least 200,000 retrieved images from the source domain. Too few source images give rise to overfitting quickly. Therefore, the initial number of retrieved nearest neighbors ($\mathcal{K}_0$) for each target training sample is set to meet this requirement. On the other hand, a surprising result is that setting $\mathcal{K}_0$ too large would make the performance of the target learning task drop significantly. In our experiments, we set $\mathcal{K}_0$ to different values for Stanford Dogs ($\mathcal{K}_0=100$), Oxford Flowers ($\mathcal{K}_0=300$), Caltech 256 ($\mathcal{K}_0=50-100$), and MIT Indoor 67 ($\mathcal{K}_0=100$). Since there exists much overlap among the nearest neighbors of different target samples, the retrieved images typically do not cover the entire ImageNet or Places datasets.

\subsection{Fine-grained Object Recognition}
\noindent\textbf{Stanford Dogs 120.} Stanford Dogs 120~\cite{khosla2011novel} contains 120 categories of dogs. There are 12000 images for training, and 8580 images for testing. We do not use the parts information during selective joint fine-tuning, and use the commonly used mean class accuracy to evaluate the performance as in \cite{griffin2007caltech}.

As shown in Table \ref{tab:StanfordDogs120}, the mean class accuracy achieved by fine-tuning the residual network using the training samples of this dataset only and without a source domain is 80.4\%.
It shows that the 152-layer residual network~\cite{he2015deep,he2016identity} pre-trained on the ImageNet dataset~\cite{russakovsky2015imagenet} has a strong generalization capability on this fine-grained classification task. Using the entire ImageNet dataset during regular joint fine-tuning can improve the performance by 5.1\%. When we finally perform our proposed selective joint fine-tuning using a subset of source domain images retrieved using histograms of low-level convolutional features, the performance is further improved to 90.2\%, which is 9.8\% higher than the performance of conventional fine-tuning without a source domain and 4.3\% higher than the result reported in \cite{krause2015unreasonable}, which expands the original target training set using Google image search. This comparison demonstrates that selective joint fine-tuning can significantly outperform conventional fine-tuning.


\begin{table}[t]\small
\setlength{\abovecaptionskip}{10pt}
\setlength{\belowcaptionskip}{-10pt}
\begin{center}
\begin{tabular}{@{}lccc@{}}
\toprule
Method                                                     & mean Acc(\%)       \\ \midrule
HAR-CNN         \cite{xie2015hyper}                        & 49.4               \\
Local Alignment \cite{gavves2015local}                     & 57.0               \\
Multi scale metric learning \cite{qian2015fine}            & 70.3               \\
MagNet   \cite{rippel2016metric}                           & 75.1               \\
Web Data + Original Data \cite{krause2015unreasonable}     & 85.9               \\\midrule
Training from scratch using target domain only             & 53.8\\
Selective joint training from scratch                      & 83.4\\
Fine-tuning w/o source domain                              & 80.4               \\
Joint fine-tuning with all source samples                  & 85.6               \\
Selective joint FT with random source samples              & 85.5               \\
Selective joint FT w/o iterative NN retrieval              & 88.3               \\
Selective joint FT with Gabor filter bank                  & 87.5               \\
Selective joint fine-tuning                                & 90.2               \\
Selective joint FT with Model Fusion                       & \textbf{90.3}      \\
\bottomrule
\end{tabular}
\end{center}
\caption{Classification results on Stanford Dogs 120.\vspace{-6mm}}
\label{tab:StanfordDogs120}
\end{table}

\begin{table}[b]\small
\setlength{\abovecaptionskip}{10pt}
\setlength{\belowcaptionskip}{-10pt}
\begin{center}
\begin{tabular}{@{}lcc@{}}
\toprule
Method                                                           & mean Acc(\%)       \\ \midrule
MPP \cite{yoo2015multi}                                          & 91.3               \\
Multi-model Feature Concat  \cite{azizpour2015generic}          & 91.3               \\
MagNet   \cite{rippel2016metric}                                 & 91.4               \\
VGG-19 + GoogleNet + AlexNet \cite{kim2015learning}              & 94.5               \\ \midrule
Training from scratch using target domain only                   & 58.2\\
Selective joint training from scratch                            & 80.6\\
Fine-tuning w/o source domain                                    & 92.3               \\
Joint fine-tuning with all source samples                       & 93.4               \\
Selective joint FT with random source samples                    & 93.2               \\
Selective joint FT w/o iterative NN retrieval                    & 94.2               \\
Selective joint FT with Gabor filter bank                        & 93.8               \\
Selective joint fine-tuning                                      & 94.7               \\
Selective joint FT with model fusion                             & \textbf{95.8}      \\ \midrule
VGG-19 + Part Constellation Model \cite{simon2015neural}         & 95.3               \\
Selective joint FT with val set                                  & \textbf{97.0}      \\
\bottomrule
\end{tabular}
\end{center}
\caption{Classification results on Oxford Flowers 102. The last two rows compare performance using the validation set as additional training data.}
\label{tab:Flowers102}
\end{table}

\noindent\textbf{Oxford Flowers 102.} Oxford Flowers 102~\cite{nilsback2008automated} consists of 102 flower categories. 1020 images are used for training, 1020 for validation, and 6149 images are used for testing. There are only 10 training images in each category.

As shown in Table \ref{tab:Flowers102}, the mean class accuracy achieved by conventional fine-tuning using the training samples of this dataset only and without a source domain is 92.3\%. Selective joint fine-tuning further improves the performance to 94.7\%, 3.3\% higher than previous best result from a single network~\cite{rippel2016metric}.
To compare with previous state-of-the-art results obtained using an ensemble of different networks, we also average the performance of multiple models obtained during iterative source image retrieval for hard training samples in the target domain. Experiments show that the performance of our ensemble model is 95.8\%, 1.3\% higher than previous best ensemble performance reported in \cite{kim2015learning}. Note that Simon \emph{et al.}~\cite{simon2015neural} used the validation set in this dataset as additional training data. To verify the effectiveness of our joint fine-tuning strategy, we have also conducted experiments using this training setting and our result from a single network outperforms that of \cite{simon2015neural} by 1.7\%.

\begin{table*}[t]\small
\centering
\begin{tabular}{@{}lcccc@{}}
\toprule
Method                     & \begin{tabular}[c]{@{}c@{}}mean Acc(\%)\\ 15/class\end{tabular} & \begin{tabular}[c]{@{}c@{}}mean Acc(\%)\\ 30/class\end{tabular} & \begin{tabular}[c]{@{}c@{}}mean Acc(\%)\\ 45/class\end{tabular} & \begin{tabular}[c]{@{}c@{}}mean Acc(\%)\\ 60/class\end{tabular} \\ \midrule
M-HMP \cite{bo2013multipath}                  & 40.5$\pm$0.4                                                       & 48.0$\pm$0.2                                                       & 51.9$\pm$0.2                                                       & 55.2$\pm$0.3                                                       \\
Z. \& F. Net  \cite{zeiler2014visualizing}                  & 65.7$\pm$0.2                                                       & 70.6$\pm$0.2                                                       & 72.7$\pm$0.4                                                       & 74.2$\pm$0.3                                                       \\
VGG-19  \cite{VGGnet}          & -                                                              & -                                                              & -                                                              & 85.1$\pm$0.3                                                       \\
VGG-19 + GoogleNet +AlexNet  \cite{kim2015learning}               & -                                                              & -                                                              & -                                                              & 86.1                                                           \\
VGG-19 + VGG-16  \cite{VGGnet}           & -                                                              & -                                                              & -                                                              & 86.2$\pm$0.3                                                       \\\midrule
Fine-tuning w/o source domain     & 76.4$\pm$0.1                                                           & 81.2$\pm$0.2                                                           & 83.5$\pm$0.2                                                           & 86.4$\pm$0.3                                                          \\
Selective joint fine-tuning & \textbf{80.5$\pm$0.3}                                                           & \textbf{83.8$\pm$0.5}                                                           & \textbf{87.0$\pm$0.1}                                                           & \textbf{89.1$\pm$0.2}                                                           \\ \bottomrule
\end{tabular}
\caption{Classification results on Caltech 256.\vspace{-6mm}}
\label{tab:Caltech256}
\end{table*}

\subsection{General Object Recognition}
\noindent\textbf{Caltech 256.} Caltech 256~\cite{griffin2007caltech} has 256 object categories and 1 background cluster class. In every category, there are at least 80 images used for training, validation and testing. Researchers typically report results with the number of training samples per class falling between 5 and 60.
We follow the testing procedure in \cite{wang2010locality} to compare with state-of-the-art results.

We conduct four experiments with the number of training samples per class set to 15, 30, 45 and 60, respectively.
According to Table~\ref{tab:Caltech256}, in comparison to conventional fine-tuning without using a source domain, selective joint fine-tuning improves classification accuracy in all four experiments, and the degree of improvement varies between 2.6\% and 4.1\%. Performance improvement due to selective joint fine-tuning is more obvious when a smaller number of target training image per class are used. This is because limited diversity in the target training data imposes a greater need to seek help from the source domain.
In most of these experiments, the classification performance of our selective joint fine-tuning is also significantly better than previous state-of-the-art results.

\begin{table}[h]\small
\setlength{\abovecaptionskip}{10pt}
\setlength{\belowcaptionskip}{-10pt}
\begin{center}
\begin{tabular}{@{}lcc@{}}
\toprule
Method                                                     & mean Acc(\%)       \\ \midrule
MetaObject-CNN \cite{wu2015harvesting}                     & 78.9               \\
MPP + DFSL \cite{yoo2015multi}                             & 80.8               \\
VGG-19 + FV \cite{cimpoi2016deep}                          & 81.0               \\
VGG-19 + GoogleNet  \cite{kim2015learning}                 & 84.7               \\
Multi scale + multi model ensemble \cite{herranz2016scene} & 86.0               \\ \midrule
Fine-tuning w/o source domain                              & 81.7               \\
Selective joint FT with ImageNet$^{\mbox{(i)}}$            & 82.8               \\
Selective joint FT with Places$^{\mbox{(ii)}}$             & 85.8               \\
Selective joint FT with hybrid data$^{\mbox{(iii)}}$       & 85.5               \\
Average the output of (ii) and (iii)                       & \textbf{86.9}      \\
\bottomrule
\end{tabular}
\end{center}
\caption{Classification results on MIT Indoor 67.\vspace{-6mm}}
\label{tab:MitIndoor67}
\end{table}

\subsection{Scene Classification}
\noindent\textbf{MIT Indoor 67.} MIT Indoor 67~\cite{quattoni2009recognizing} has 67 scene categories. In each category, there are 80 images for training and 20 images for testing. Since MIT Indoor 67 is a scene dataset, in addition to the ImageNet ILSVRC 2012 training set~\cite{russakovsky2015imagenet}, the Places-205 training set~\cite{zhou2014learning} is also a potential source domain. We compare three settings during slective joint fine-tuning: ImageNet as the source domain, Places as the source domain, and the combination of both ImageNet and Places as the source domain.

As shown in Table~\ref{tab:MitIndoor67}, the mean class accuracy of selective joint fine-tuning with ImageNet as the source domain is 82.8\%, 1.1\% higher than that of conventional fine-tuning without using a source domain. Since ImageNet is an object-centric dataset while MIT Indoor 67 is a scene dataset, it is hard for training images in the target domain to retrieve source domain images with similar low-level characteristics. But source images retrieved from ImageNet still prevent the network from overfitting too heavily and help achieve a performance gain. When the Places dataset serves as the source domain, the mean class accuracy reaches 85.8\%, which is 4.1\% higher than the performance of fine-tuning without a source domain and 4.8\% higher than previous best result from a single network~\cite{cimpoi2016deep}. And the hybrid source domain based on both ImageNet and Places does not further improve the performance. Once averaging the output from the networks jointly fine-tuned with Places and the hybrid source domain, we obtain a classification accuracy 0.9\% higher than previous best result from an ensemble model~\cite{herranz2016scene}.

\subsection{Ablation Study}
\vspace{-0mm}
We perform an ablation study on both Stanford Dogs 120 \cite{khosla2011novel} and Oxford Flowers 102 \cite{nilsback2008automated} by replacing or removing a single component from our pipeline. First, instead of fine-tuning, we perform training from scratch in two settings, one using the target domain only and the other using selective joint training. Tables~\ref{tab:StanfordDogs120} and \ref{tab:Flowers102} show that while selective joint training obviously improves the performance, it is still inferior than fine-tuning pretrained networks. This is because we only subsample a relatively small percentage (20-30\%) of the source data, which is still insufficient to train deep networks from scratch. Second, instead of using a subset of retrieved training images from the source domain, we simply use all training images in the source domain. Joint fine-tuning with the entire source domain decrease the performance by 4.6\% and 1.3\% respectively. This demonstrates that using more training data from the source domain is not always better. On the contrary, using less but more relevant data from the source domain is actually more helpful. Third, instead of using a subset of retrieved training images, we use the same number of randomly chosen training images from the source domain. Again, the performance drops by 4.7\% and 1.5\% respectively. Fourth, to validate the effectiveness of iteratively increasing the number of retrieved images for hard training samples in the target domain, we turn off this feature and only use the same number ($\mathcal{K}_0$) of retrieved images for all training samples in the target domain. The performance drops by 1.9\% and 0.5\% respectively. This indicates that our adaptive scheme for hard samples is useful in improving the performance. Fifth, we use convolutional kernels in the two bottom layers of a pre-trained AlexNet as our filter bank. If we replace this filter bank with the Gabor filter bank, the overall performance drops by 2.7\% and 0.9\% respectively, which indicates a filter bank learned from a diverse dataset could be more powerful than an analytically defined one. Finally, if we perform conventional fine-tuning without using a source domain, the performance drop becomes quite significant and reaches 9.8\% and 2.4\% respectively.
\vspace{-2mm}

\section{Conclusions}
\vspace{-2mm}
In this paper, we address deep learning tasks with insufficient training data by introducing a new deep transfer learning scheme called selective joint fine-tuning, which performs a target learning task with insufficient training data simultaneously with another source learning task with abundant training data. Different from previous work which directly adds extra training data to the target learning task, our scheme borrows samples from a large-scale labeled dataset for the source learning task, and do not require additional labeling effort beyond the existing datasets. Experiments show that our deep transfer learning scheme achieves state-of-the-art performance on multiple visual classification tasks with insufficient training data for deep networks. Nevertheless, how to find the most suitable source domain for a specific target learning task remains an open problem for future investigation.
\vspace{-4mm}

\paragraph{Acknowledgment}
This work was partially supported by Hong Kong Innovation and Technology Fund (ITP/055/14LP).

{\small
\bibliographystyle{ieee}
\bibliography{egbib}
}

\end{document}